\pdfoutput=1

\documentclass[11pt]{article}

\usepackage[]{acl}

\usepackage{times}
\usepackage{latexsym}

\usepackage[T1]{fontenc}

\usepackage[utf8]{inputenc}

\usepackage{microtype}
\usepackage{graphicx}

\usepackage{CJK}
\usepackage{indentfirst}
\usepackage{amsmath}
\usepackage{cases}
\usepackage{subfigure}
\usepackage{multirow}

\usepackage{inconsolata}

\usepackage{graphicx}
\usepackage{tabularx}
\usepackage{multirow}
\usepackage{booktabs}

\title{Knowledge Boundary and Persona Dynamic \\ Shape A Better Social Media Agent}

\author{
Junkai Zhou$^{1,2}$,
Liang Pang$^{1}$\thanks{\ \ Corresponding authors},
Ya Jing$^{1,2}$\thanks{\ \ Equal Contributions},
Jia Gu$^{1,2\dagger}$, 
Huawei Shen$^{1,2}$,
Xueqi Cheng$^{1,2}$\\
$^{1}$CAS Key Laboratory of AI Security, \\
Institute of Computing Technology, Chinese Academy of Sciences, Beijing, China \\
 $^{2}$University of Chinese Academy of Sciences, Beijing, China \\
{\tt\ \{zhoujunkai20z,pangliang,shenhuawei,cxq\}@ict.ac.cn}}

\begin{document}
\maketitle
\begin{abstract}
Constructing personalized and anthropomorphic agents holds significant importance in the simulation of social networks.
However, there are still two key problems in existing works: the agent possesses world knowledge that does not belong to its personas, and it cannot eliminate the interference of diverse persona information on current actions, which reduces the personalization and anthropomorphism of the agent.
To solve the above problems, we construct the social media agent based on personalized knowledge and dynamic persona information.
For personalized knowledge, we add external knowledge sources and match them with the persona information of agents, thereby giving the agent personalized world knowledge.
For dynamic persona information, we use current action information to internally retrieve the persona information of the agent, thereby reducing the interference of diverse persona information on the current action.
To make the agent suitable for social media, we design five basic modules for it: persona, planning, action, memory and reflection. To provide an interaction and verification environment for the agent, we build a social media simulation sandbox.
In the experimental verification, automatic and human evaluations demonstrated the effectiveness of the agent we constructed. The code will be released on \url{https://github.com/934865517zjk/Social-agent}.
\end{abstract}

\section{Introduction}

\begin{figure}
    \centering 
    \includegraphics[width=7.8cm]{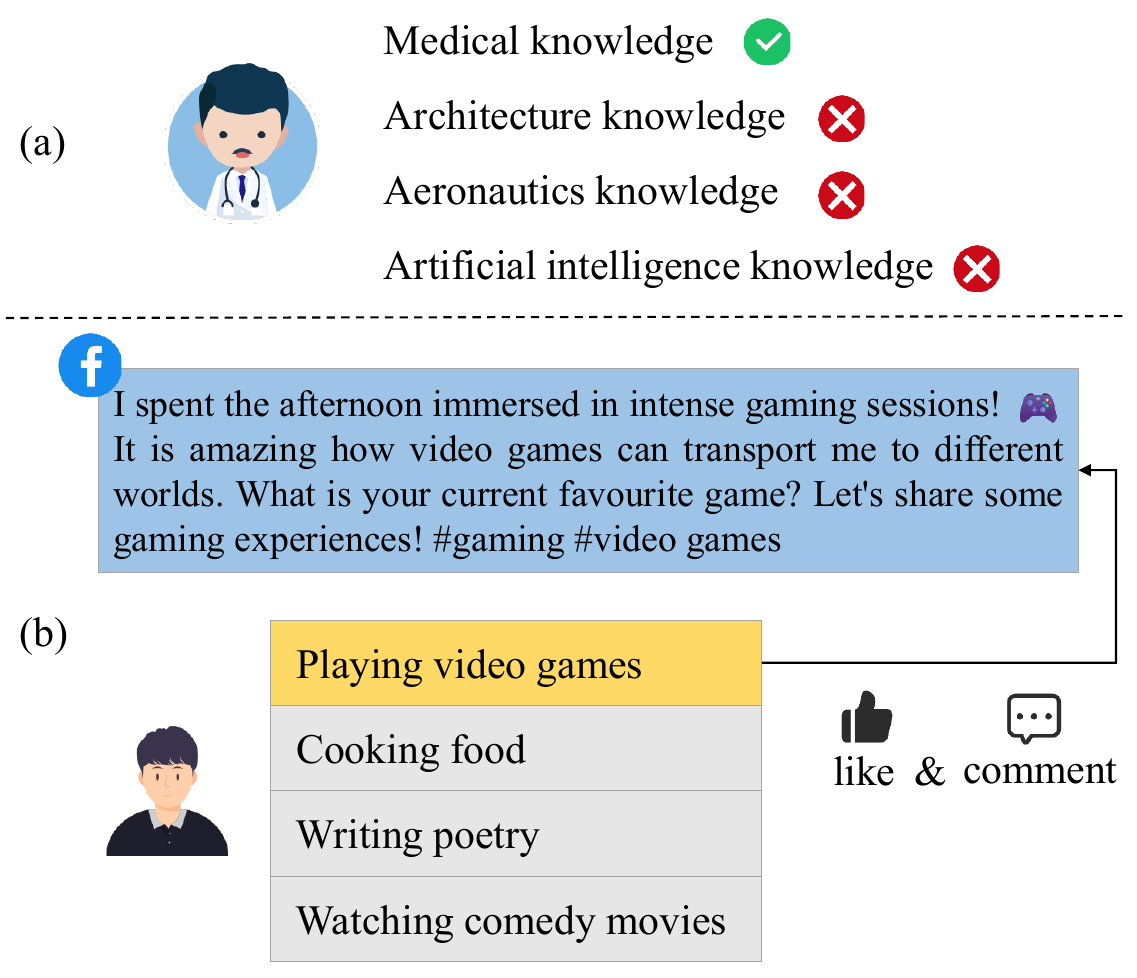}
    \caption{
    Knowledge boundary and persona dynamic shape personalization in social media: (a) knowledge boundary: the agent should only have world knowledge that is consistent with the persona; (b) persona dynamic: the agent should only use personal information related to the current action when performing actions.
    }
    \label{fig:fig1}
\end{figure}

Social networks play a crucial role in the daily lives of individuals \citep{bakshy2012role, forestier2012roles}. In sociology, the social network is an important way to understand behavior \citep{jackson2014networks, thomas2017understanding}. 
By studying the performance of the agent in social networks, we can better understand its behavior and guide the construction of it.
The emergence of large language models (LLMs) \citep{openai2022chatgpt, openai2023gpt4, touvron2023llama, touvron2023llama2} brings possibilities for the agent construction. Several works attempt at it and achieve encouraging results \citep{park2023generative, li2023camel, chen2023agentverse}. Due to the social nature of social media, it is particularly important to shape the personalization of the agent, which facilitates deep and meaningful interactions with other users. To balance the information from knowledge and persona, we shape personalization through two important attributes: the knowledge boundary and the persona dynamic. However, existing works ignore these two key problems, which make the performance of the agent different from humans.

Regarding the knowledge boundary, the agent should only process world knowledge that aligns with the given persona. However, the knowledge boundary of existing works is too flexible~\cite{park2023generative, xiao2023simulating}, and they use all knowledge within LLMs or directly retrieve it from knowledge datasets. This makes the agent more like a ``naturalist'' who frequently outputs knowledge that does not belong to its persona, rather than a personalized agent. It may reduce its personalization and credibility and cause users to lose trust. As shown in Figure~\ref{fig:fig1}(a), the doctor should have professional medical knowledge and not professional knowledge of law, physics, and artificial intelligence.
Therefore, the knowledge possessed by each agent should be a limited subset related to persona, rather than the entire world knowledge.

For persona dynamic, the agent should only use relevant persona information for the current action. However, existing works~\cite{wang2023large, qian2023communicative} are less flexible in using persona information. They combine all the persona information with the action information and provide them to the agent when it takes action. The persona information is usually diverse, but the current action may only be related to a subset of it.
As shown in Figure~\ref{fig:fig1}(b), the persona information contains multiple hobbies, and the current action is only related to ``playing video games'' in the hobby. 
If all persona information is used, possible conflicts between them may interfere with the decision-making of the action, or the agent may generate generic text that covers all persona information but contains little specific content. Therefore, the agent should only use the persona information relevant to the action rather than all of them.

To address the above challenges, we build a social media agent based on personalized knowledge and dynamic persona information. For personalized knowledge, we add retrievable external knowledge sources to the agent and combine them with persona information. Every time the agent uses knowledge, it needs to judge whether the knowledge the persona should have. If so then the agent uses the knowledge, otherwise it discards the knowledge. The basis for judgment is whether the text similarity between this knowledge and the description of the knowledge that the persona should have is greater than a fixed threshold.
For dynamic persona information, we divide the persona information into multiple retrievable items, including detailed historical behavior information, social media content preferences, and persona knowledge. Each retrieval item is about a single historical behavior, content preference, or knowledge. When the agent performs an action, the action information is used as a query, and the above three are retrieved respectively to obtain the relevant persona information, thereby reducing the interference of other irrelevant persona information.

To give the agent basic functions and make it suitable for social media, we equip the agent with five modules: persona, action, planning, memory, and reflection. In the persona module, the agent is given rich persona information and personalized world knowledge. In the action module, the agent can complete various actions on social media, such as likes and posts. The planning module provides the activity and planning basis for the action module. In the memory module, the historical action information of the agent is recorded, and the agent can retrieve the memory when needed to support actions. In the reflection module, the agent can reflect on past actions and support more complex actions such as following users.

To provide an interaction and verification environment for agents, we build a social media simulation sandbox based on an open-source platform. The sandbox has the basic functions of social media, and the agent can complete actions by calling the API. Meanwhile, the recommendation mechanism is added to the sandbox to improve the realism. Overall user activity is simulated and combined with the planning module of the agent to make the sandbox closer to social media in the real world.

In the experiment, multiple agents interact in the sandbox and take actions such as browsing, liking and posting. In automated evaluation, the posts engaged by the agent through liking, reblogging, or commenting exhibit higher entailment or textual similarity with persona information compared to posts browsed only. 
The posts and comments generated by the agent have good text quality and are relevant to the persona information.
The evaluation based on GPT-4 also confirms the above results. In human evaluation, the action rationality and text generation quality of the agent are evaluated, and the results show the effectiveness of the agent.

\begin{figure*}
    \centering 
    \includegraphics[width=15.5cm]{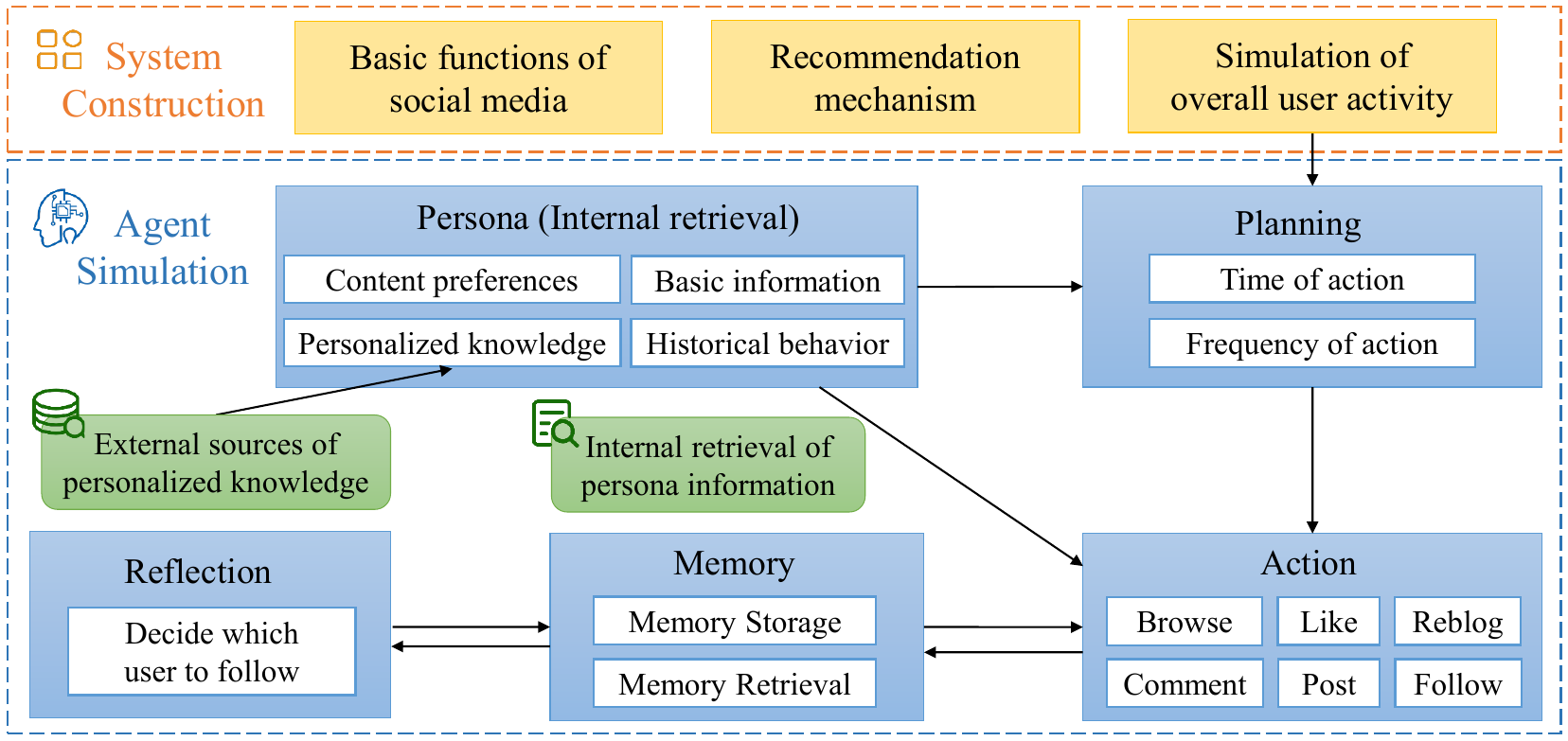}
    \caption{
    The overall framework of our work is divided into two parts: system construction and agent simulation. The system construction part consists of the basic functions of social media, the recommendation mechanism and the simulation of overall user activity. The agent simulation part consists of five modules: persona, planning, action, memory and reflection. In the persona module, the agent can obtain personalized knowledge from external knowledge sources, and the persona information is internally retrievable when performing actions.
    }
    \label{fig:fig2}
\end{figure*}

Our contributions in this paper are three folds:
\begin{itemize}
    \item We endow the agent with personalized world knowledge and dynamic persona information, thereby improving the personalization and anthropomorphism of the agent.
    \item We design five modules for the agent: persona, action, planning, memory and reflection, which make it more suitable for social media.
    \item We build a social media simulation sandbox with authentic page interfaces, offering the agent an interaction and verification environment. Automatic and human evaluations demonstrate the effectiveness of the agent.
\end{itemize}

\section{Related Work}
After the emergence of large language models, recent works have used them in the research of agents, aiming to get closer to general artificial intelligence.
Part of these works focus on building a single agent and improving the performance of the agent on various tasks \citep{shen2023hugginggpt, shinn2023reflexion, wang2023voyager, pallagani2024prospects}.
Other works focus on the research of multi-agent collaboration, which allows each agent to play different roles to cooperate and complete more complex tasks \citep{li2023camel, qian2023communicative, chen2023agentverse, hong2023metagpt}.

The performance of agents in social environments is an issue that needs urgent research, and several works have made preliminary attempts at this.
\citet{park2023generative} build a social simulation sandbox to observe whether the action of agents in it is close to real people.
\citet{xiao2023simulating} simulate the performance of multiple agents when facing public management crisis events. 
Other works explore the interaction of agents in social networks to observe the performance of agents in the virtual world.
\citet{park2022social} develop a simulation sandbox to refine the design of social media platforms by examining edge actions and implementing interventions.
\citet{gao2023s} simulate and observe the attitude of agents facing specific events by introducing real social media data.

\section{Our Approach}
The overall framework of our work is divided into two parts: agent simulation and system construction, which is shown in Figure~\ref{fig:fig2}.

\begin{figure*}
    \centering 
    \includegraphics[width=15.7cm]{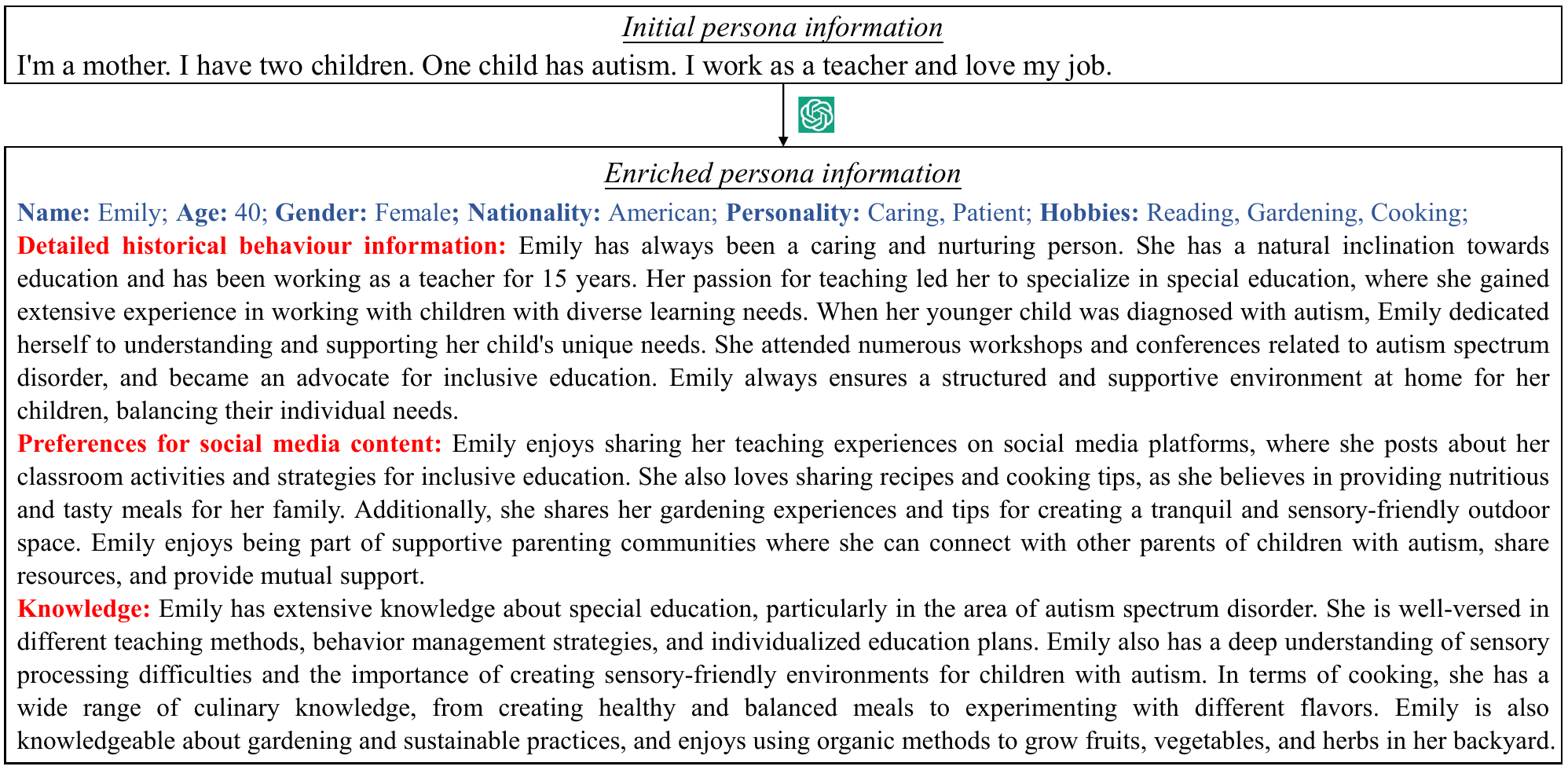}
    \caption{
    An example of the process of enriching persona information. The text marked in blue is the basic information of the persona. The text marked in red is the advanced attributes of the persona.
    }
    \label{fig:fig3}
\end{figure*}

\subsection{Agent Simulation}
We design five modules for the agent to make it suitable for social media: persona, planning, action, memory and reflection, which follows as \citet{park2023generative} and \citet{wang2023large}.\footnote{The relationship between the five modules of our agent and existing works is discussed in Appendix~\ref{sec:appendix_H}.}

\subsubsection{Persona Module}
\paragraph{Creation and Enrichment of Persona}
We enrich the simple persona in Personachat \citep{zhang-etal-2018-personalizing} and create the rich persona information of agents. Personachat is a persona-based dialogue dataset containing 1155 personas. The enriched persona information includes basic information and advanced attributes. Basic information includes name, age, gender, nationality, personality and hobbies. Advanced attributes include detailed historical behavior information, social media content preferences, and persona knowledge. An example of the process of enriching persona information is shown in Figure~\ref{fig:fig3}. We design the prompt for ChatGPT to complete the above process. The implementation details are shown in Appendix~\ref{sec:appendix_A}.

\paragraph{Addition of Personalized Knowledge}
Each agent should have knowledge related to the persona instead of all world knowledge, so we need to endow each persona with personalized knowledge. We use the HotpotQA \citep{yang2018hotpotqa} as an external knowledge source, boasting 113k question-answer pairs and encompassing extensive world knowledge.
When the agent generates a post, it retrieves knowledge that the persona should have from HotpotQA based on the post topic, thus providing support for the action.
Specifically, we use the post topic as a query and use ColBERT \citep{santhanam2022colbertv2} to retrieve the relevant knowledge from HotpotQA.
The TF-IDF algorithm \cite{salton1988term} is then used to calculate the text similarity between the retrieved knowledge and the ``knowledge'' item in persona information. When the text similarity between the two is greater than the knowledge adoption threshold $T_{k}$, the knowledge is used, otherwise, it is not used.

\paragraph{Dynamic Persona Information}
Since the information of each persona is diverse, which includes multiple hobbies, historical behavior information, social media content preferences, etc.
Excessive persona information may prevent the agent from selecting the most relevant persona information for the current action, thereby interfering with the action decision-making process.
Therefore, we set up the advanced attributes of persona as multiple internal retrieval items, including detailed historical behavior information, social media content preferences, and persona knowledge.
When the agent actions, the action information is used as a query and each advanced attribute is retrieved internally. The persona information that is most relevant to the current action in each advanced attribute is obtained and used to support the action, thereby avoiding the interference of irrelevant persona information.

\begin{figure*}
    \centering 
    \includegraphics[width=14.7cm]{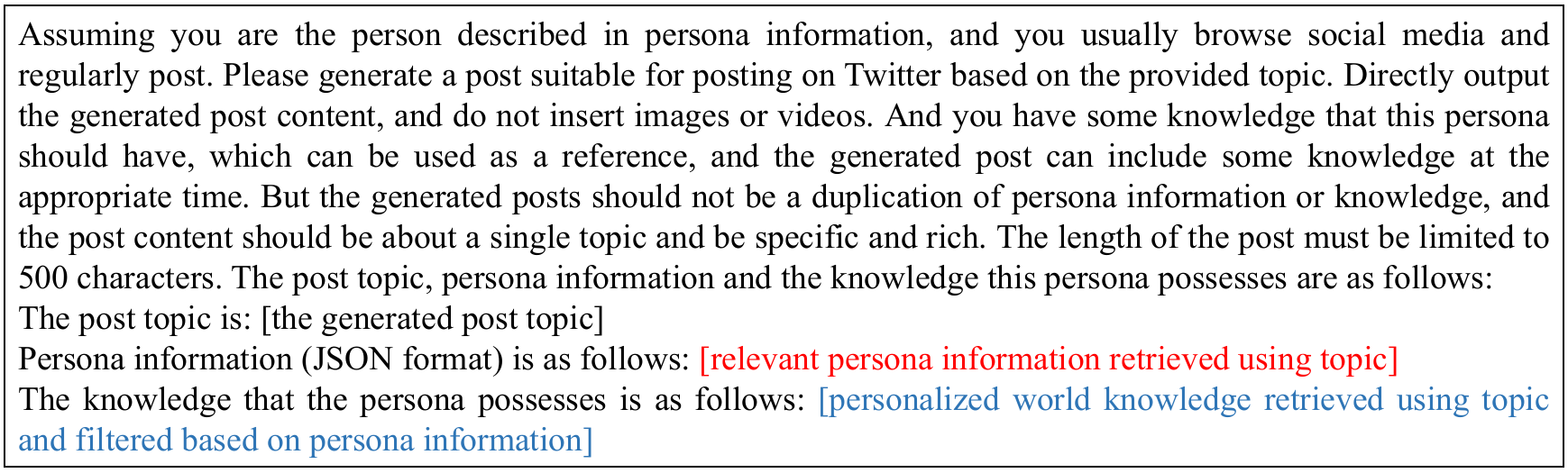}
    \caption{
    The prompt for the agent to generate post content. Text marked in red is personalized knowledge retrieved using post topics. The text marked in blue is the persona information obtained through internal retrieval.
    }
    \label{fig:fig_post}
\end{figure*}

\subsubsection{Action Module}
In the action module, agents can perform actions on social media.
Our agent is specifically constructed for social media, so we only simulate actions suitable for it. We design six actions: browse, like, reblog, comment, post and follow users. 
The description of them is as follows: (1) Browse: the agent looks through posts on social media. Since we use LLMs instead of multi-modal models, the post content only includes text but not images and videos. (2) Like: the agent expresses its approval or liking of the content through the like operation. (3) Reblog: the agent shares posts of other people on its homepage. (4) Comment: the agent leaves responses on a post, fostering conversation with the content creator. (5) Post: the agent creates content for others to browse. (6) Follow users: the agent subscribes to the homepages of other users.

We technically use prompt engineering \citep{brown2020language} to implement the above actions. In the designed prompt, we splice together the instructions for the agent to make action decisions, the retrieved persona information and the action information. 
Due to the complexity of the posting action, personalized external world knowledge is also added to the above prompt to better support the agent in generating posts. 
For likes, reblogs and comments, we use the browsed post content as a query to internally retrieve the persona information of the agent and obtain the relevant persona information. For posting, the agent first generates post topics based on all persona information, and internally retrieves the persona information based on each topic. In addition, the topic is used as a query to retrieve world knowledge, and the text similarity between the knowledge and the ``knowledge'' item in persona information is calculated. If the text similarity is greater than the knowledge adoption threshold $T_{k}$, the knowledge will be used, otherwise it will not be used. The prompt for the agent to generate post content is shown in Figure~\ref{fig:fig_post}. More prompts of the action are shown in Appendix~\ref{sec:appendix_B}.

\subsubsection{Planning Module}
In the planning module, the agent can plan its action. The agent generates the plan containing the time and probabilities of the action based on its own user activity and persona information, which includes the time to browse posts every day, the time and number to post every week, the probability of likes, reblogs and comments. The agent performs actions according to the generated plan. Meanwhile, we convert the action probability into the number of actions and make the agent follow the constraints of it. An example of planning content is shown in Figure~\ref{fig:fig4}. The prompt for generating planning content is shown in Appendix~\ref{sec:appendix_C}.

\begin{figure}
    \centering 
    \includegraphics[width=6.8cm]{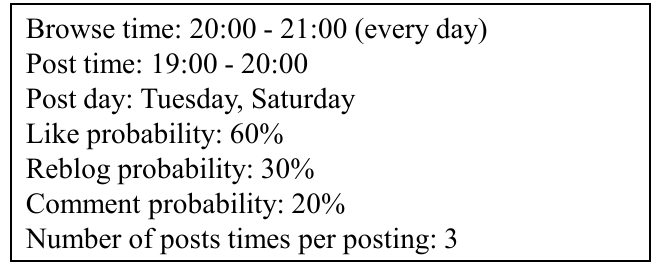}
    \caption{
    An example of the generated planning content.
    }
    \label{fig:fig4}
\end{figure}

\subsubsection{Memory Module}
\paragraph{Memory Storage}
We equip the agent with short-term memory and long-term memory. Information about the actions of the agent is recorded and stored as short-term memory. Taking likes as an example, the recorded action information includes the post content, whether to like it, the poster ID, etc. The external knowledge source of persona serves as the long-term memory of agents, which comes from the HotpotQA dataset \citep{yang2018hotpotqa}.

\paragraph{Memory Retrieval}
In memory retrieval, the agent can extract information related to the current action from short-term memory and long-term memory to better support the action.
Specifically, when the agent generates a post, it uses the post topic as a query to retrieve the long-term memory and obtain relevant knowledge. 
After the post is generated, the short-term memory is retrieved and the similarity between the generated post and its own previous posts is calculated.
If the similarity is greater than the post duplication threshold $T_{p}$, the agent regenerates the post to avoid publishing posts that are too repetitive. The above two text similarities are calculated through the TF-IDF algorithm \cite{salton1988term}.

\subsubsection{Reflection Module}
Based on short-term memory, the agent can generate high-level ideas through reflection. Every time the simulation time increases by two days, the agent reflects on the users to follow based on its history action information including likes, reblogs, and comments. 
We design the appropriate prompt for the agent and take its history action information as input. The agent outputs the user ID to follow or not to follow.
Because of the length of the post on social media, we equip the agent with a summary component. The summarized post and action information of the agent are spliced and input into the reflection module, thereby improving the effect of the reflection module. The prompts for generating the reflection and the summary are shown in Appendix~\ref{sec:appendix_D} and Appendix~\ref{sec:appendix_E} respectively.
                
\subsection{System Construction}
We construct a social media simulation sandbox, which mainly includes three parts: basic functions of the social media, recommendation mechanism and overall user activity simulation.
\subsubsection{Construction of Social Media Platform}
We construct a social media sandbox with authentic page interfaces based on Mastodon.\footnote{\url{https://github.com/tootsuite/mastodon}} Mastodon is an open-source social media software similar to Twitter that provides basic social media functions such as live feeds and posting. The screenshot of the sandbox we constructed is shown in Appendix~\ref{sec:appendix_F}. To allow the agent to automatically execute actions in the sandbox, we encapsulate the existing Mastodon API\footnote{\url{https://github.com/halcy/Mastodon.py}} to build an API interface, which has a variety of functions: register accounts, get live feeds, post, like, comment, etc. In addition, we design timers to simulate real-world time. The simulation sandbox runs according to turns, and the time increases by one hour for each turn.

\subsubsection{Recommendation Mechanism}
To make our sandbox closer to real social media, we add a recommendation mechanism to it. We build our recommendation module based on the Mastodon Digest\footnote{\url{https://github.com/hodgesmr/mastodon_digest}}, which is a recommendation project for Mastodon.
Every time a user browses the posts, it scores and sorts unread posts. The score is calculated as follows.
\begin{equation}\label{eq1}
Score=\sqrt[3]{N_{L} \cdot N_{R} \cdot N_{C}} / \sqrt{N_{F}},
\end{equation}
where $N_{L}$, $N_{R}$, and $N_{C}$ are the number of likes, reblogs, and comments of the post, respectively. $N_{F}$ is the number of followers of the post creator.

\subsubsection{Simulation of Overall User Activity}
In real social media, different users may have different levels of activity. We simulate the overall activity of users. Following as \citet{wang2023large}, user activity usually presents a long-tail distribution\footnote{\url{https://en.wikipedia.org/wiki/Long_tail}} from an overall perspective, that is, a few users are highly active and most users are low active. 
Therefore, the Pareto distribution\footnote{\url{https://en.wikipedia.org/wiki/Pareto_distribution}} is used to simulate user activity, and the probability density of user activity is as follows.
\begin{equation}\label{eq2}
p(x)=\alpha x_{min}^{\alpha} / (\alpha+1),
\end{equation}
where $x_{min}$ is the minimum activity level, and $\alpha$ is the parameter that affects the shape of the function.
After the user activity is generated, we combine it with the planning module of the agent. Specifically, user activity and persona information are input into the plan generation process, thereby jointly determining the action probability in the plan.

\begin{table*}[htp]
\centering
\scalebox{0.86}{
\begin{tabular}{lccccccc}
\toprule
&  \textbf{\multirow{2}{*}{Action}} & \multicolumn{4}{c}{\textbf{Automatic}}  & & \multicolumn{1}{c}{\textbf{Human}} \\
\cline{3-6} \cline{8-8}
\textbf{} & &  BERTScore$\uparrow$ & C.score$\uparrow$ & Distinct-1$\uparrow$ & Distinct-2$\uparrow$ & & Rationality$\uparrow$ \\
\hline
\textbf{\multirow{7}{*}{First Stage}} & not like & 0.433 & 0.934 & \multirow{2}{*}{-} & \multirow{2}{*}{-} &  & \multirow{2}{*}{0.678} \\
& like & \textbf{0.440} & \textbf{0.972} & & & & \\
\cline{2-8}
 & not reblog & 0.432 & 0.928 & \multirow{2}{*}{-} & \multirow{2}{*}{-} &  &  \multirow{2}{*}{0.656} \\
 & reblog & \textbf{0.438} & \textbf{0.967} &  &  & &\\
 \cline{2-8}
 & not comment & 0.433 & \textbf{0.946} & - & - &  & \multirow{2}{*}{0.544}  \\
 & comment & \textbf{0.435} & 0.944 & 3.38 & 19.05 & &  \\
 \cline{2-8}
& post & 0.465  & 0.973 & 22.40 & 63.88 & & - \\
\hline

\textbf{\multirow{7}{*}{Second Stage}} &  not like & 0.428 & 0.931 & \multirow{2}{*}{-} & \multirow{2}{*}{-} & & \multirow{2}{*}{0.778} \\
& like & \textbf{0.435} & \textbf{0.953} &  &  & & \\
\cline{2-8}
& not reblog & 0.428  & 0.937 & \multirow{2}{*}{-} & \multirow{2}{*}{-} & & \multirow{2}{*}{0.622} \\
 & reblog & \textbf{0.434}  & \textbf{0.942} &  &  & &  \\
 \cline{2-8}
 & not comment & 0.428  & \textbf{0.962} & - & - & &  \multirow{2}{*}{0.477}  \\
 & comment & \textbf{0.431}  & 0.939 & 2.98 & 17.17 & &   \\
 \cline{2-8}
 & Post & 0.466  & 0.983 & 22.24 & 64.10 & & - \\
\bottomrule
\end{tabular}
}
\caption{\label{table1}
Automatic evaluation and human evaluation results of agent actions.}
\end{table*}

\begin{table}[htp]
\centering
\scalebox{0.88}{
\begin{tabular}{lccccc}
\toprule
\textbf{} & & \textbf{Info.$\uparrow$}  & \textbf{Cons.$\uparrow$}  & \textbf{Hum.$\uparrow$} \\
\hline
& Comment (1st)  & 1.767 & 2.292 & 2.950 \\
& Comment (2nd) & 1.983 & 2.317 & 2.983 \\
\hline
& Post (1st)  & 2.658 & 2.967 & 3.000 \\ 
& Post (2nd)  & 2.533 & 2.983 & 3.000  \\
\bottomrule
\end{tabular}
}
\caption{\label{table2}
Text generation quality evaluation results based on GPT-4. 1st and 2nd represent the first stage of interaction and the second stage of interaction respectively.}
\end{table}

\section{Experiments}
\label{sec:expen}
\subsection{Experiment Settings}
\paragraph{Models and Implementation Details}
We use ChatGPT \citep{openai2022chatgpt} as the basic model for constructing the agent. We use gpt-3.5-turbo provided from the API of OpenAI.\footnote{\url{https://openai.com/api/}} 
Parameters are set as follows: the knowledge adoption threshold $T_{k}$ is 0.25, the post duplication threshold $T_{p}$ is 0.80, and $\alpha$ in the Pareto distribution is 2, determined through parameter tuning. The agent count is 450, with 150 initial agents and 300 regular agents. The sandbox starts empty, with the initial agent providing initial content and publishing 1050 posts. Regular agents subsequently engage in two stages of interaction within the sandbox. In the first stage, they post and browse the posts posted by initial agents. New posts posted by the regular agent in this stage will not be browsed. In the second stage, regular agents post and browse posts from regular agents. Each stage is one week in simulation time.

\subsection{Evaluation Metrics}
\paragraph{Automatic Metrics}
We choose the automatic metrics in different aspects to evaluate different actions of agents. The action should be consistent with the persona information, we evaluate the relationship between the posts that the agent chooses to like, forward, or comment on and the persona information. The relationship between posts generated by the agent and persona information is also evaluated. We use consistency score (\textbf{C.score}) \citep{madotto-etal-2019-personalizing} and BERTScore \citep{zhang2019bertscore} to evaluate consistency and text similarity respectively. For C.score, the RoBERTa \citep{DBLP:journals/corr/abs-1907-11692} fine-tuned on DialogueNLI \citep{welleck-etal-2019-dialogue} as the natural language inference (NLI) model is used to evaluate the consistency between persona information and post content. When the relation between them is entailment, neutral, and contradiction, the C.score is 1, 0, and -1, respectively.


For posts and comments, we additionally evaluate the text generation quality of the agent. We use \textbf{distinct-1/2} \citep{li-etal-2016-diversity} to evaluate the diversity. In addition, we randomly select 120 posts and 120 comments and use GPT-4 \citep{openai2023gpt4} to score. The version we use is gpt-4-1106-preview.
Informativeness (\textbf{Info}), consistency (\textbf{Cons}), and humanness (\textbf{Hum}) are used to evaluate text quality. Informativeness means whether the text is diverse and informative. Consistency means whether the text is relevant and consistent with the persona information. Humanness means how similar the generated text is to text written by a real person. The three indicators are scored on a scale of 1 to 3, where 3 is good, 2 is moderate, and 1 is poor. When evaluating, only personality, hobbies, and social media content references of the persona are used, because the action is more related to the above three attributes, and there is a limit on the input length of the BERT, RoBERTa and GPT-4.

\paragraph{Human Evaluations}
In human evaluation, three graduate students with good English skills are asked to evaluate the action rationality and text generation quality of the agent. The action rationality is 0 or 1, 1 indicates that the action is reasonable, 0 otherwise. For the evaluation of the reflection module, the annotator determines whether the user the agent follows is reasonable based on the historical actions of the agent. If it is reasonable, the reflection score (\textbf{Ref}) is 1, 0 otherwise. 
The metrics used in text quality are the same as those used in the evaluation based on GPT-4. In addition, the appropriateness of knowledge (\textbf{Know}) is evaluated. If the agent utilizes knowledge when needed and uses it appropriately, Know is 1; if the agent does not use it when knowledge is unnecessary, it is also 1. Otherwise, it is 0.
To measure the agreement between human annotators, we use Pearson correlation coefficient \citep{cohen2009pearson}. We randomly select 30 regular agents for the action rationality evaluation. For text quality, we randomly select 60 posts and 60 comments posted by regular agents.

\begin{table}[htp]
\centering
\scalebox{0.85}{
\begin{tabular}{lcccc}
\toprule
\textbf{} &  \textbf{Info.$\uparrow$}  & \textbf{Cons.$\uparrow$}  & \textbf{Hum.$\uparrow$}  & \textbf{Know.$\uparrow$} \\
\hline
Comment (1st) & 2.422 & 1.922 & 2.533 & - \\
Comment (2nd) & 2.522 & 2.011 & 2.600 & - \\
\hline
Post (1st)  & 2.733 & 2.911 & 2.600 & 0.889  \\
Post (2nd)  & 2.589 & 2.944 & 2.667 & 0.778  \\
\bottomrule
\end{tabular}
}
\caption{\label{table3}
Human evaluation results of text generation quality. 1st and 2nd represent the first stage of interaction and the second stage of interaction respectively.}
\end{table}

\subsection{Results}
\paragraph{Results of Action Rationality}
As shown in Table~\ref{table1}, the posts engaged with by the agent through liking, reblogging, or commenting exhibit higher BERTScore with persona information compared to posts browsed only. 
The consistency between these engaged posts and person information is greatly high, exceeding browse-only posts except for the comment.
The above results show that the agent can make reasonable actions based on its persona. In the human evaluation of the action rationality, the like and forward achieve good scores. The rationality of comments is lower than the above two, which is because comments are more complex and difficult for the agent to decide. The agent shows similar results in two stages of interactions, indicating that the agent can interact well with users possessing different persona information.

\paragraph{Results of Text Generation Quality}
In automatic evaluation, posts generated by the agent show good diversity and informativeness, both using automatic metrics and GPT-4-based evaluations. The comments generated by the agent are less diverse and informative. This is because comments are shorter and tend not to include much content. As shown in Table~\ref{table1}, the posts generated by the agent have good BERTScore and consistency score with the persona information, indicating that the agent can generate posts that are relevant and consistent with the persona. In the evaluation based on GPT-4, there is good consistency between the generated posts and the persona information, as shown in Table~\ref{table2}. The consistency between the generated comments and persona information is lower than that in posts. We believe this is because generating comments does not involve too much persona information. In addition, the generated posts and comments have great humanness, indicating that they are similar to those written by real people.

\begin{figure}[htp]
    \centering 
    \includegraphics[width=7cm]{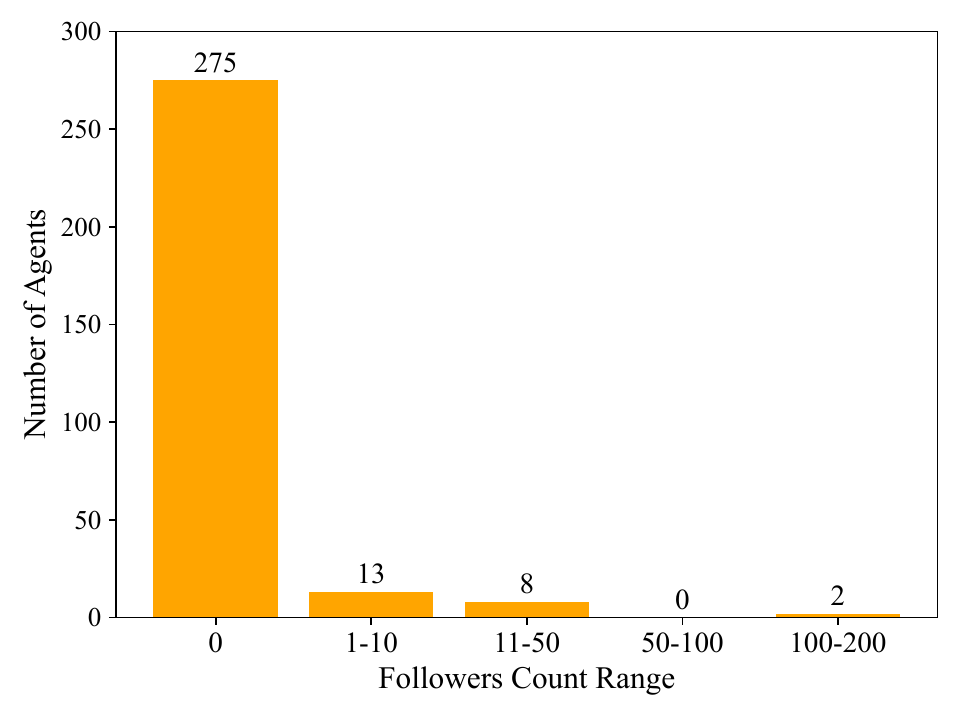}
    \caption{The number of followers of the agent.}
    \label{fig:fig6}
\end{figure}

The human evaluation shows similar results. As shown in Table~\ref{table3}, the posts and comments have good informativeness, with posts being more informative due to their length. The posts have higher consistency than comments because they contain more information related to the persona. Both posts and comments show good humanness, indicating that annotators prefer that they are written by real people. The knowledge score is high, indicating that the agent can select and use knowledge appropriately. The Pearson correlation coefficients of annotators for the action and text quality are 0.40 and 0.79 respectively, indicating good consistency between annotators. The action exhibits a lower consistency score compared to text quality due to its subjectivity in evaluation. We provide the results of the ablation study, examples of action and text generation in Appendix~\ref{sec:appendix_G0} and Appendix~\ref{sec:appendix_G} respectively.

\paragraph{Results of Reflection}
We manually evaluate the effect of the reflection module, and the reflection score is 0.644, which shows that the agent can reasonably choose the users to follow based on its historical action information.
In addition, we count the number of followers of 300 regular agents, and the results are shown in Figure~\ref{fig:fig6}. The results show that most agents have few or no followers, while two agents have more than 100 followers, forming well-known influences. This is similar to real social media, illustrating the effectiveness of the agent and simulation sandbox we constructed.

\section{Conclusion}
In this work, we construct the social media agent and a social media simulation sandbox. We endow the agent with personalized world knowledge and dynamic persona information, thereby improving its personalization and anthropomorphism. Five modules are designed to make the agent suitable for social media: persona, action, planning, memory and reflection. Meanwhile, we build a sandbox to provide the agent with an interaction and verification environment. Experimental results demonstrate the effectiveness of our constructed agent.

\section*{Limitations}
The construction of agents is an important step towards general artificial intelligence. We make attempts and construct more personalized and anthropomorphic agents by giving them personalized world knowledge and dynamic persona information. However, there are still some limitations in our work and things that need to be improved in future work. The following are the main two points:

The content in the social media sandbox is all text-modal. When constructing agents, we use LLMs like ChatGPT instead of multi-modal models. LLMs can only input text content, but cannot input images, audio, video, etc. Therefore, posts and comments in our social media sandbox contain text-only content. This is somewhat different from real social media. In future work, we will explore agent construction based on multi-modal models and add image and video content to the social media sandbox.

There are some differences between the personalized knowledge possessed by agents and users in the real world. When giving the agent personalized knowledge, we use the world knowledge from the HotpotQA dataset and combine it with the persona information of the agent, thereby endowing the agent with personalized world knowledge. Although we simulate that each user has different world knowledge in this way, there are still differences between users in the real world. For example, all the knowledge sources we use come from the HotpotQA dataset, but real users may have other world knowledge. In future work, we will consider more diversified sources of knowledge and assign them to agents.

\section*{Ethics Statement}
In this work, we use LLMs to conduct research, which involves generating text. Therefore, we have the same concerns as other LLMs and text generation research. For example, there is a risk of generating toxic or biased language. 
When constructing the agent, it is necessary to give the agent persona information. The purpose of using persona information is to improve the personalization of the agent rather than guessing user identities. The Personachat used in this work is a public dataset, so there are no issues such as privacy concerns. When using the Personachat dataset and project resources such as Mastodon and Mastodon Digest, we comply with the terms of use of these resources.

\section*{Acknowledgements}
This work was supported by the National Key R\&D Program of China (2022YFB3103700, 2022YFB3103704), the National Natural Science Foundation of China (NSFC) under Grants No. 62276248, U21B2046, and the Youth Innovation Promotion Association CAS under Grant No. 2023111.

\bibliography{anthology,custom}

\clearpage

\appendix

\section{Prompts for Enriching Persona}
\label{sec:appendix_A}
The prompt designed for enriching persona information is shown in Table~\ref{table_A1}. By using it, we can enrich the simple persona information from the Personachat dataset into diverse and rich persona information. 

\section{Prompts for Agent Actions}
\label{sec:appendix_B}
The prompt designed to determine whether the agent likes a post is shown in Table~\ref{table_B1}. By using it, we can simulate the liking action of the agent and generate decision information on whether to like based on persona information and post content.
The prompt designed to determine whether the agent reblogs a post is shown in Table~\ref{table_B2}. By using it, we can simulate the reblogging action of the agent and generate decision information on whether to reblog based on persona information and post content.
The prompt designed to determine whether the agent comments on a post is shown in Table~\ref{table_B3}. By using it, we can simulate the commenting action of the agent and generate decision information on whether to comment or comment on content based on persona information and post content.

The prompt designed for generating topics for posts is shown in Table~\ref{table_B4}. By using it, the agent can generate the required number of post topics. The prompts designed for agents to generate posts based on world knowledge are shown in Table~\ref{table_B5}. By using it, the agent can generate posts based on a given topic, persona information, and the provided world knowledge. The prompts designed for agents to post without using world knowledge are shown in Table~\ref{table_B6}. By using it, the agent can generate posts based on a given topic and persona information.

\section{Prompts for Planning Module}
\label{sec:appendix_C}
The prompt designed for generating the plan of the agent is shown in Table~\ref{table_C1}. By using it, the agent can generate an action based on its own user activity and persona information, including the time and probability of action.

\section{Prompts for Reflection Module}
\label{sec:appendix_D}
The prompt designed to follow users is shown in Table~\ref{table_D1}. By using it, the agent can determine whether to follow users based on their historical action information. The prompts designed to concatenate historical action information of the agent are shown in Table~\ref{table_D2}.

\section{Prompts for Summary Component}
\label{sec:appendix_E}
The prompt designed to summary the content of the post is shown in Table~\ref{table_E1}. Since the concatenated text during reflection is too long, we can summarize the content of the post by using it.

\section{The Screenshot of Our Social Media Simulation Sandbox}
\label{sec:appendix_F}
\begin{figure*}
    \centering 
    \includegraphics[width=15cm]{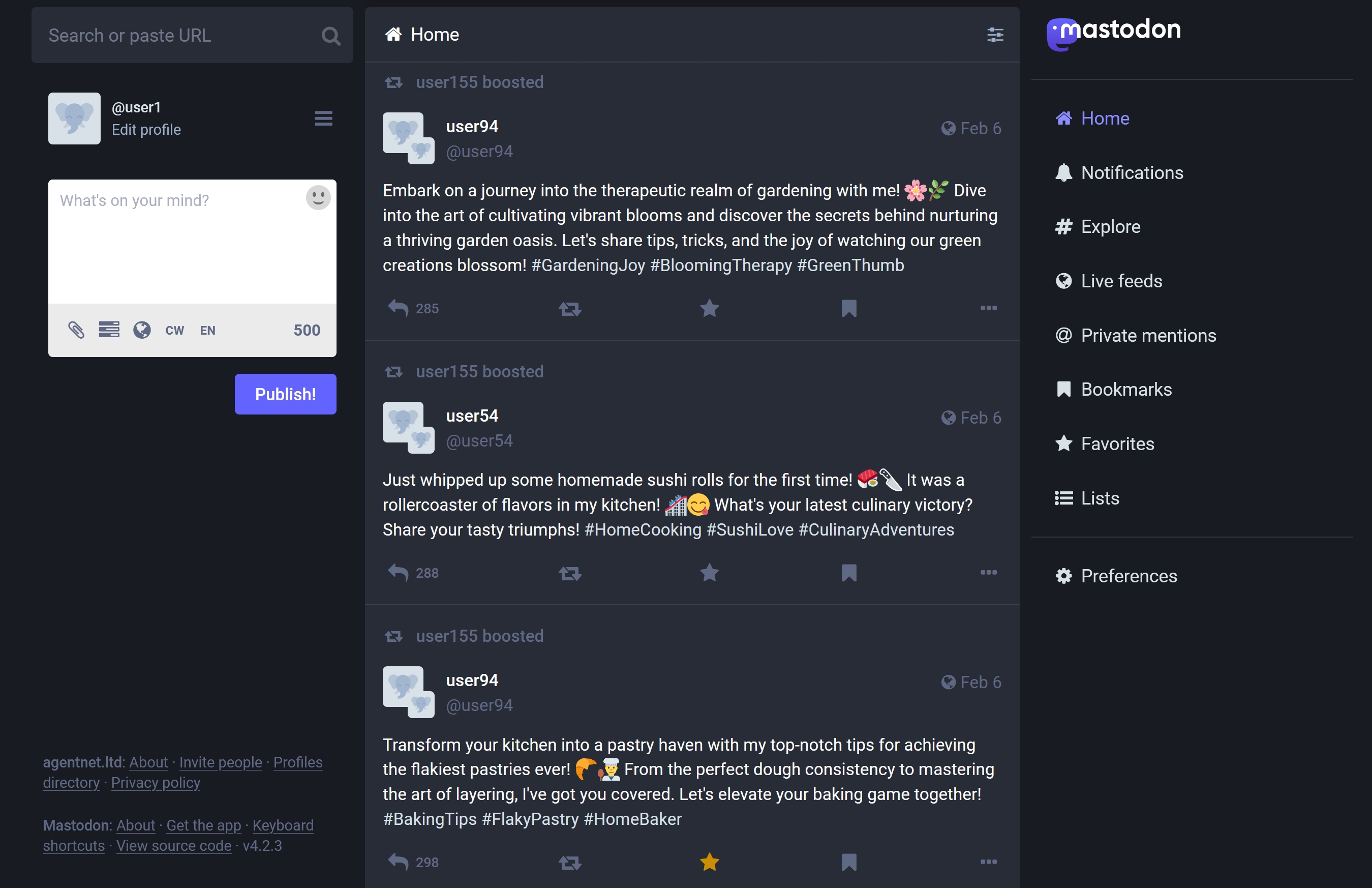}
    \caption{
    Screenshot of the home page of our social media simulation sandbox.
    }
    \label{fig:home_small}
\end{figure*}

\begin{figure*}
    \centering 
    \includegraphics[width=15cm]{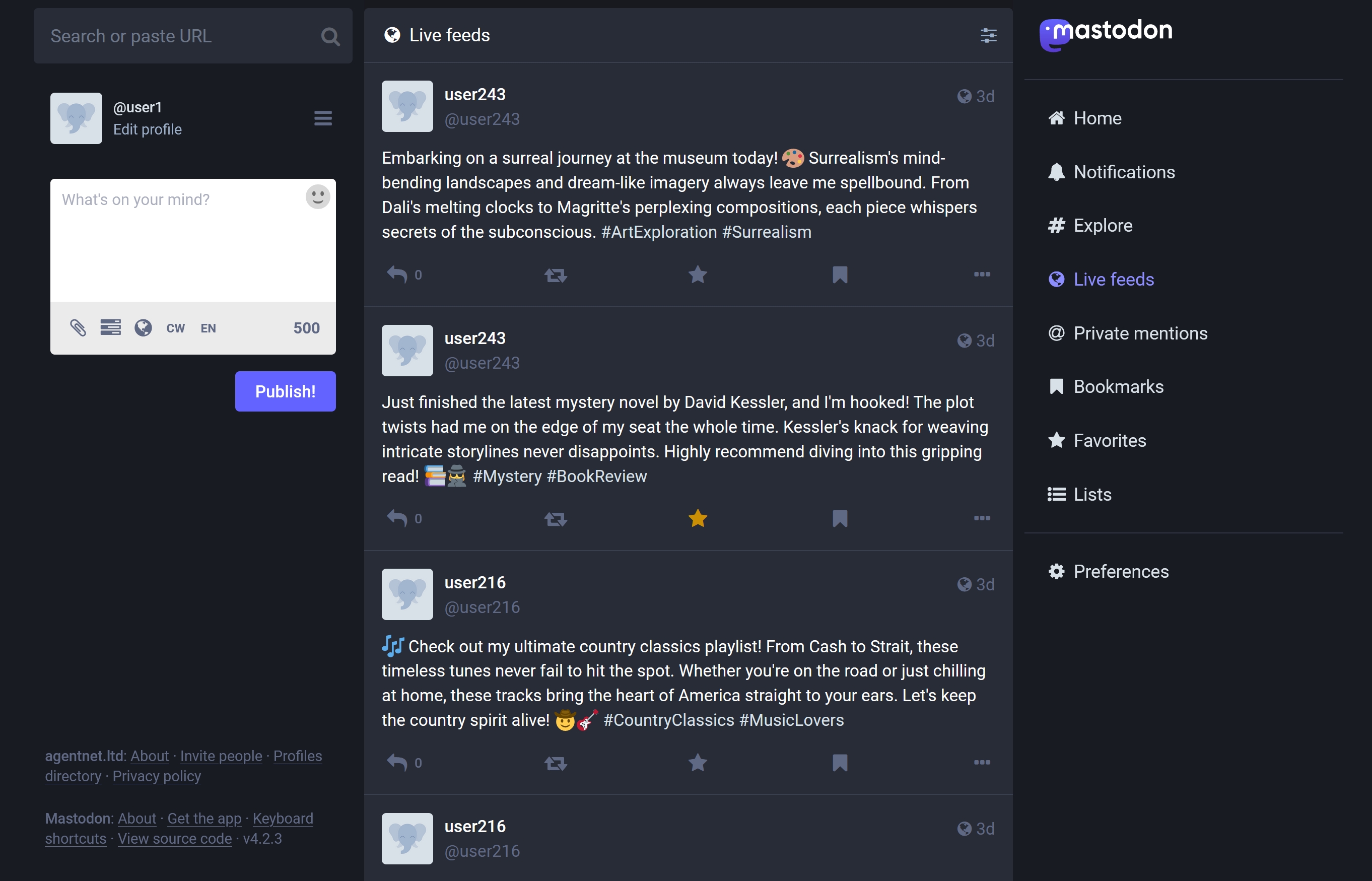}
    \caption{
    Screenshot of the live feeds page of our social media simulation sandbox.
    }
    \label{fig:live_feed}
\end{figure*}

\begin{figure*}
    \centering 
    \includegraphics[width=15cm]{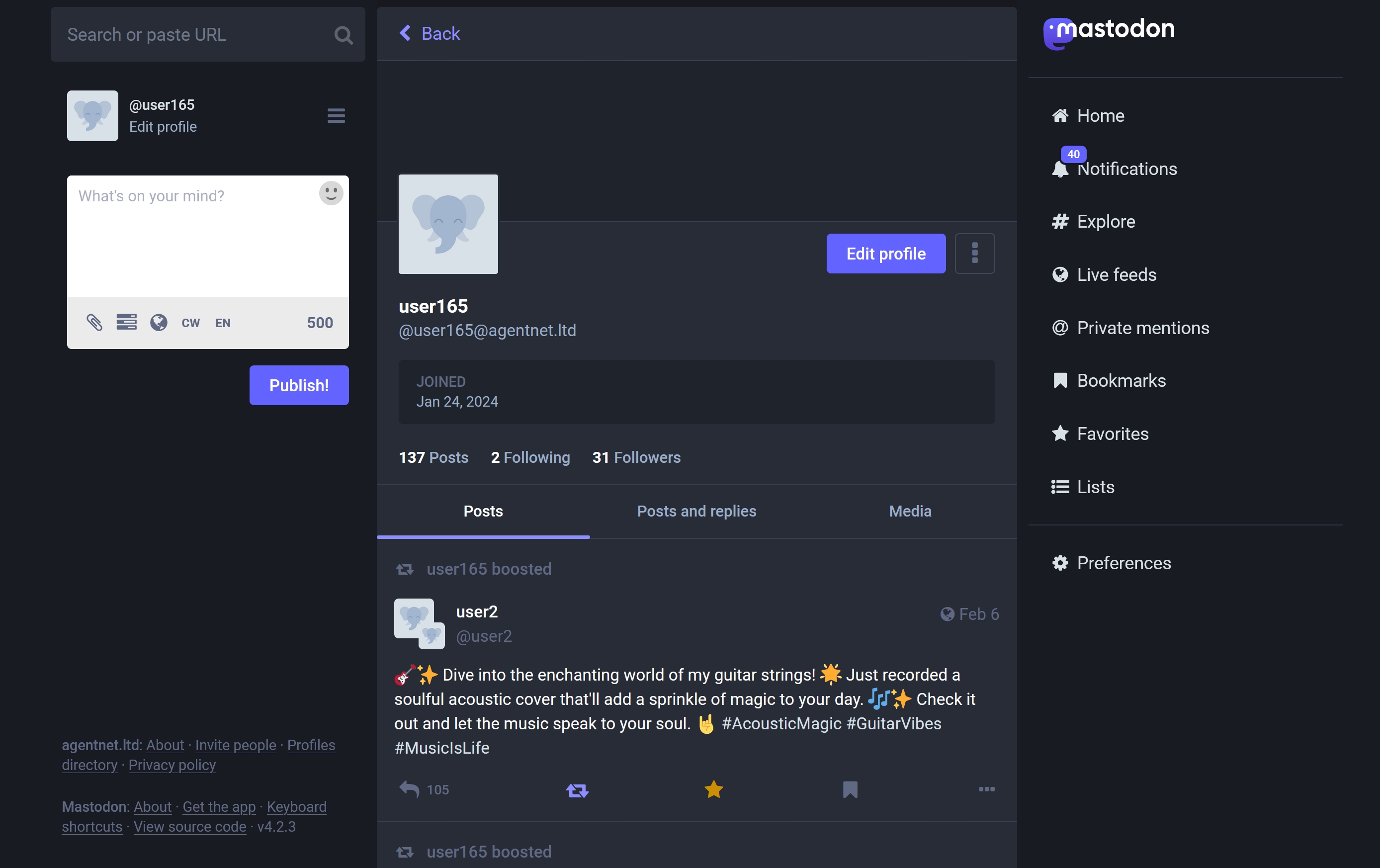}
    \caption{
    Screenshot of user profile page from our social media simulation sandbox.
    }
    \label{fig:main_home}
\end{figure*}

\begin{figure*}
    \centering 
    \includegraphics[width=15cm]{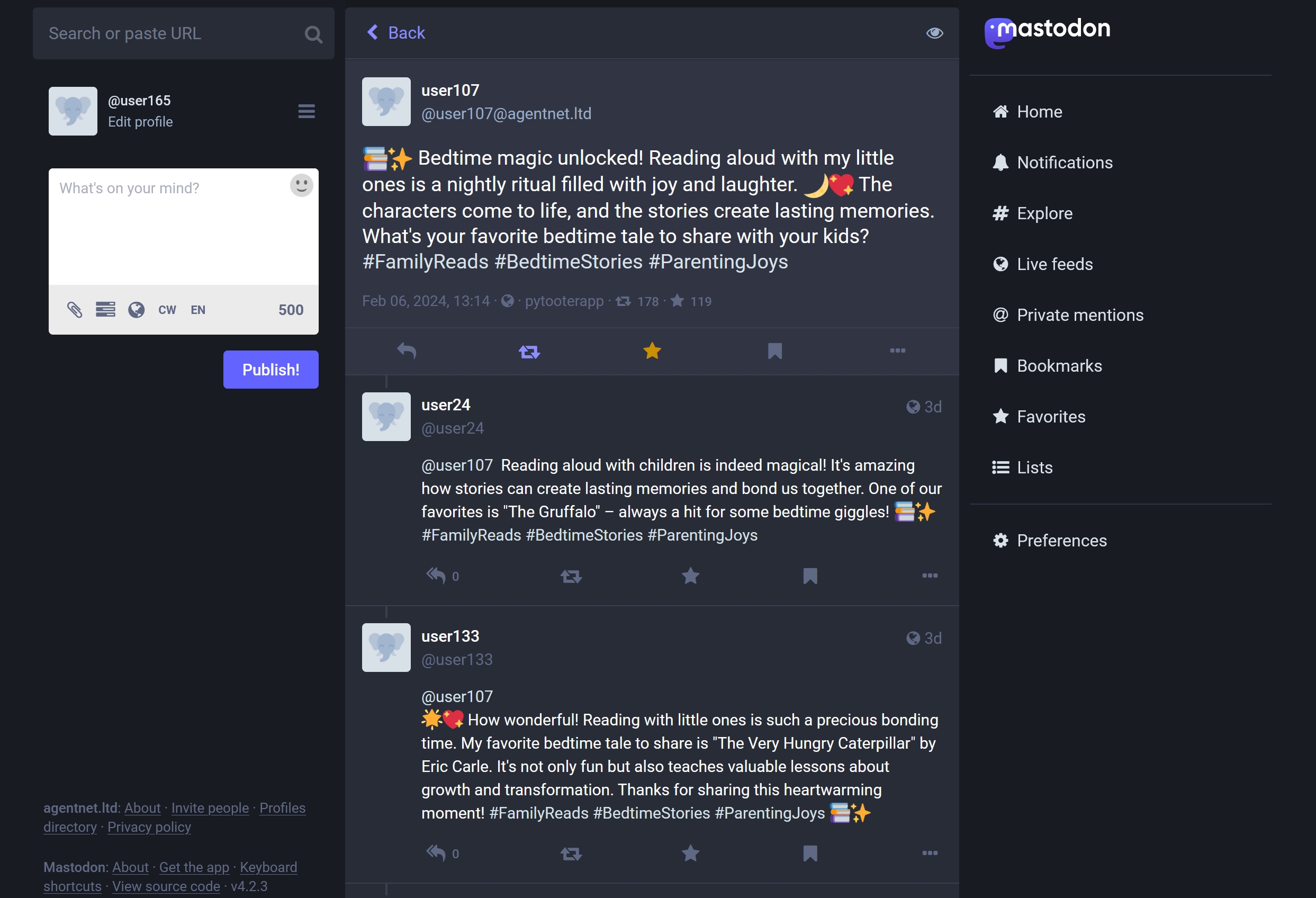}
    \caption{
    Screenshot of a post from our social media simulation sandbox.
    }
    \label{fig:post_example}
\end{figure*}

\begin{figure*}
    \centering 
    \includegraphics[width=15cm]{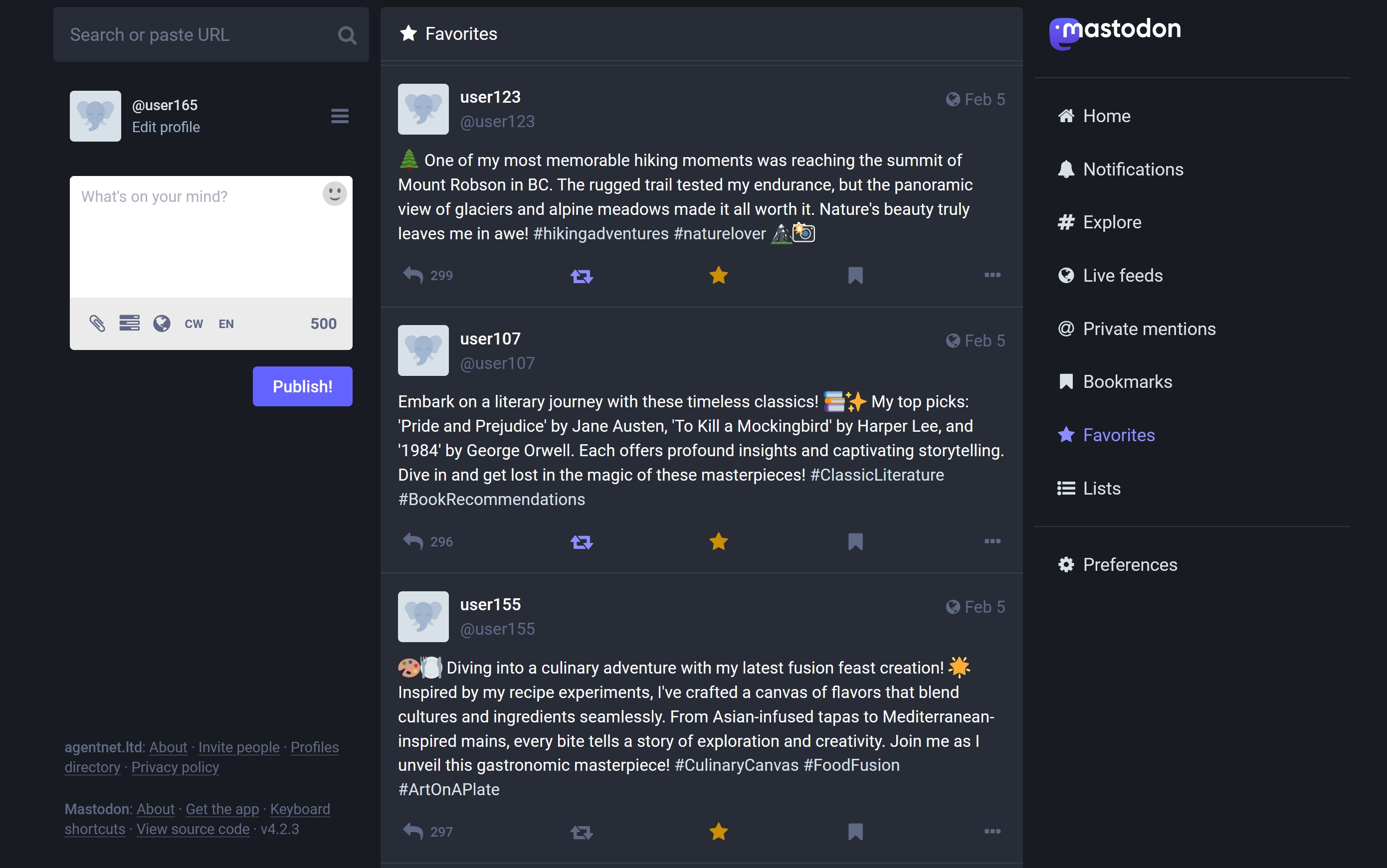}
    \caption{
    Screenshot of favorite post list of a user in our social media simulation sandbox.
    }
    \label{fig:favorite}
\end{figure*}

\begin{figure*}
    \centering 
    \includegraphics[width=15cm]{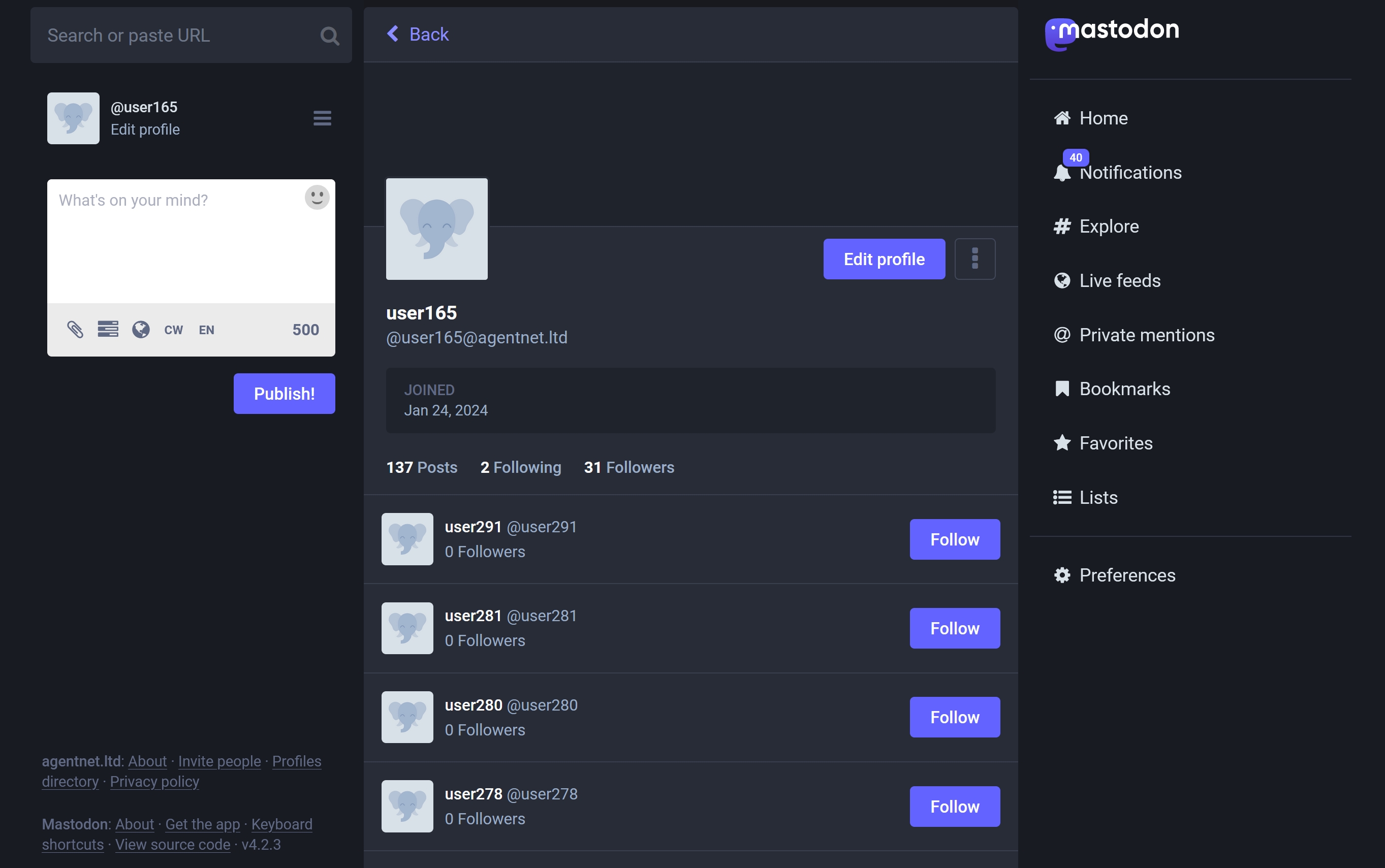}
    \caption{
    Screenshot of follower list of a user in our social media simulation sandbox.
    }
    \label{fig:main_home_follower}
\end{figure*}

We provide part of screenshots of pages from our social media simulation sandbox. The home page displays recommended posts and the latest updates from following users, as shown in Figure~\ref{fig:home_small}. The live feeds page displays the latest updates from all users, as shown in Figure~\ref{fig:live_feed}. The user profile page displays the personal information of the user, published and reblogged posts, follow functions, etc, as shown in Figure~\ref{fig:main_home}. In Figure~\ref{fig:post_example}, we provide a screenshot of a post. In Figure~\ref{fig:favorite}, we provide a screenshot of the like post list of a user. In Figure~\ref{fig:main_home_follower}, we provide a screenshot of the follower list of a user.
 
\section{Ablation Study}
\label{sec:appendix_G0}
To verify the effectiveness of knowledge boundary (KB) and persona dynamic (PD), we conduct ablation experiments. In addition to the automatic and human evaluation indicators used in the main text, we add two automatic indicators $\Delta BS$ and $\Delta C$ to show the experimental results more intuitively. $\Delta BS$ refers to the subtraction value of BERTScore when the agent takes a certain action and BERTScore when the agent does not take a certain action. $\Delta C$ refers to the subtraction value of C.score when the agent takes a certain action and C.score when the agent does not take a certain action. The higher values of $\Delta BS$ and $\Delta C$ indicate that the action of the agent is more reasonable.

As shown in Table~\ref{table-Ab1}, when the agent uses all the persona information instead of the dynamic persona information, the automatic and human evaluation scores of likes and reblogs mostly drop, indicating that using dynamic persona information can help the agent make better decisions about likes and reblogging actions. For commenting actions, using all the persona information performs better in automatic metrics, and using dynamic persona information performs better in human evaluation. We believe this is due to the difficulty and complexity of the commenting action. For the following action, when all persona information is used, the reflection score reflecting the rationality of the following action is 0.589, which is lower than the rationality score of 0.644 using dynamic persona information, indicating that using dynamic persona information can enable the agent to reflect better.

In the text quality evaluation, when using all persona information or non-personalized knowledge, the distinct-1/2 of the posts generated by the agent drops as shown in Table~\ref{table-Ab1}, proving that their diversity becomes worse. This illustrates the necessity of dynamic persona information and personalized knowledge.
In human evaluation, when using all persona information and non-personalized knowledge, the informativeness, consistency, and humanness scores are reduced as shown in Table~\ref{table-Ab2}, which also illustrates the necessity of dynamic persona information and personalized knowledge. When using all persona information, the appropriateness of knowledge of the posting action is higher or the same as when using dynamic persona information. The Pearson correlation coefficients between annotators for actions using full persona information, text quality generated using full persona information, and non-personalized knowledge are 0.39, 0.73, and 0.68, respectively, indicating good agreement between annotators and are similar to the Pearson correlation coefficient in previous experiments.

\section{Case Study}
\label{sec:appendix_G}
We provide an example of the post action of the agent in Table~\ref{tableG}. In this example, the agent retrieves persona information and world knowledge related to ``dog behavior and effective training technique''. In the retrieved persona information, the ``detailed historical behavior information'' item mentions that she enjoys reading about dog behavior and training techniques, and the ``preferences for social media content'' item mentions that she is open to recommendations and discussions about dog training and books about animal behavior, which all help the agent generate posts on the above topics. Meanwhile, the agent obtains a piece of knowledge about a book called ``The Dog Whisperer'' by retrieving world knowledge, which is consistent with the knowledge that the persona described in the ``knowledge'' item in the persona information should have is consistent, so this knowledge is adopted. Based on them, the agent generates a relevant post, which embodies the retrieved knowledge.

The examples of like, reblog, and comment actions are shown in Table~\ref{tableG2}. In this example, the ``preferences for social media content'' item in the retrieved persona information is ``she hopes to inspire others to use their hobbies and passions to make a positive impact.'' The post that the agent browses at this time is another user passionately sharing his hobbies, which is consistent with the description in the content preferences in the above persona information. Hence, the agent chooses to like, forward and comment on this post. Meanwhile, the agent commented, ``while I'm usually more focused on animal welfare, it's fascinating to see the dedication and skill these fighters bring to the ring.'' This both reflects its persona information and compliments the content of the post.

\section{Relationship Between the Five Modules of Our Agent and Existing Works}
\label{sec:appendix_H}
The agent we constructed has five modules: persona, planning, action, memory and reflection. Among them, the persona module is following \citet{wang2023large}, and the planning, action, memory and reflection modules are following \citet{park2023generative}. It should be noted that although the concepts of these modules are derived from these two works, their modules cannot be directly used for the social media agent. To make the agent suitable for social media, we make extensive improvements, including the role of each module in the agent, design ideas and implementation methods, etc.

\clearpage

\begin{table*}[htp]
\begin{tabularx}{1.0\textwidth} {
  >{\hsize=1\hsize\linewidth=\hsize}X
}
  \hline
Please enrich the initial persona information provided by the user, including name, age, gender, nationality, personality, and hobbies. Note that this information needs to be logically consistent with the initial persona information provided by the user, and age and nationality should not always be used the same. Meanwhile, detailed historical behavior information, preferences for social media content, and knowledge should be generated. It should be as detailed as possible to help users build a virtual social media user persona with more depth and personality. 

Initial persona information provided by the user: \textbf{[Initial persona information]}

The output format is JSON format, where the keys are ``name'', ``age'', ``gender'', and ``nationality'', ``personality'', ``hobbies'', ``detailed historical behaviour information'', ``preferences for social media content'', ``knowledge''. 
The following is an output example, please strictly follow the JSON format in the example for output.

\{

\textbf{"name"}: "John",

\textbf{"age"}: 35, 

\textbf{"gender"}: "Male",

\textbf{"nationality}": "American",

\textbf{"personality"}: "Adventurous, Outgoing",

\textbf{"hobbies"}: "Working on vintage cars, Listening to country music, Taking care of dogs",

\textbf{"detailed historical behaviour information"}: "John has always been passionate about cars, especially vintage cars. He has been collecting and restoring them for the past 10 years. His love for vintage cars led him to become knowledgeable about their mechanics and history. He has also participated in local car shows and won a few awards for his beautifully restored Mustangs. John\'s dogs are his loyal companions, and he spends quality time training and playing with them. He believes in responsible pet ownership and often volunteers at local animal shelters.", 

\textbf{"preferences for social media content"}: "John enjoys sharing his car restoration projects on social media platforms, where he documents the progress and showcases the before and after pictures of his vintage Mustangs. He also loves sharing his favorite country music playlists and recommendations. Additionally, he posts adorable pictures of his dogs, sometimes showcasing their tricks and training achievements.", 

\textbf{"knowledge"}: "John has extensive knowledge about vintage cars, particularly Ford Mustangs. He is familiar with various car models, their features, and the history of the Mustang brand. He keeps up with the latest trends in car restoration techniques and actively follows vintage car communities online. In terms of country music, John has a wide knowledge of classic and contemporary country artists, their discographies, and the stories behind their songs. He also has a good understanding of training techniques and dog behavior, thanks to his experience with his two dogs." 

\}
\\
  \hline
\end{tabularx}
\caption{\label{table_A1}
The designed prompt for enriching persona information. The word in square brackets is a piece of initial persona information from the Personachat dataset that needs to be entered. The bold font in curly brackets is the key in the generated json file.
}
\end{table*}

\begin{table*}[htp]
\begin{tabularx}{1.0\textwidth} {
  >{\hsize=1\hsize\linewidth=\hsize}X
}
  \hline
Assume that you are the person described in [Persona information] and you are browsing social media. Please decide whether to like or take no action based on the content of the posts you see. When outputting, please strictly output one of ``like'' or ``no operation'' and do not output other content. The post content and persona information are as follows:

Post content: \textbf{[poster]}

Persona information: \textbf{[persona]}
\\
  \hline
\end{tabularx}
\caption{\label{table_B1}
The designed prompt for the like action. The word ``poster'' in square brackets represents the content of the post, and the word ``persona'' in square brackets represents the persona information of the agent.
}
\end{table*}

\begin{table*}[htp]
\begin{tabularx}{1.0\textwidth} {
  >{\hsize=1\hsize\linewidth=\hsize}X
}
  \hline
Assume that you are the person described in [Persona information] and you are browsing social media. Please decide whether to forward based on the content of the posts you see. When outputting, please strictly output one of ``forward'' or ``no operation'' and do not output other content. The post content and persona information are as follows:

Post content: \textbf{[poster]}

Persona information: \textbf{[persona]}
\\
  \hline
\end{tabularx}
\caption{\label{table_B2}
The designed prompt for the reblogging action. The word ``poster'' in square brackets represents the content of the post that needs to be entered, and the word ``persona'' in square brackets represents the persona information of the agent.
}
\end{table*}

\begin{table*}[htp]
\begin{tabularx}{1.0\textwidth} {
  >{\hsize=1\hsize\linewidth=\hsize}X
}
  \hline
Assume that you are the person described in [Persona information] and you are browsing social media. Please decide whether to comment based on the content of the posts you see. Note that users generally only comment on content that interests them or when they want to express their opinions. If you choose not to comment, directly output ``no comment'' and do not output other content. If you choose to comment, output the comment content directly, do not output other content, and start with ``Comment content:''. The post content and persona information are as follows: 

Post content: \textbf{[poster]} 

Persona information: \textbf{[persona]}
\\
  \hline
\end{tabularx}
\caption{\label{table_B3}
The designed prompt for the comment action. The word ``poster'' in square brackets represents the content of the post that needs to be entered, and the word ``persona'' in square brackets represents the persona information of the agent.
}
\end{table*}

\begin{table*}[htp]
\begin{tabularx}{1.0\textwidth} {
  >{\hsize=1\hsize\linewidth=\hsize}X
}
  \hline
Assume that you are the person described in [Persona information]. You usually browse social media and post regularly. Please generate \textbf{[topic count]} post topics suitable for posting on social media Twitter. The generated post topics need to be diverse and consistent with the persona information, be within 15 words in length, and do not include the topic symbol ``\#''. 

The output format is:

1. Theme one

2. Theme two

3. Theme three

The persona information is as follows: \textbf{[persona]}
\\
  \hline
\end{tabularx}
\caption{\label{table_B4}
The designed prompt for generating topics of the post. The word ``topic count'' in square brackets represents the content of the post that needs to be input, and the word ``persona'' in square brackets represents the persona information of the agent.
}
\end{table*}

\begin{table*}[htp]
\begin{tabularx}{1.0\textwidth} {
  >{\hsize=1\hsize\linewidth=\hsize}X
}
  \hline
Assume you are the person described in [Persona information], and you usually browse social media and regularly post. Please generate a post suitable for posting on Twitter based on the provided topic. Directly output the generated post content, and do not insert images or videos. And you have some knowledge that this persona should have, which can be used as a reference, and the generated post can include some knowledge at the appropriate time. But the generated posts should not be a duplication of persona information or knowledge, and the post content should be about a single topic and be specific and rich. The length of the post must be limited to 500 characters. The post topic, persona information and the knowledge this persona possesses are as follows:

The post topic is: \textbf{[topic output]}

Persona information (JSON format) is as follows: \textbf{[persona]}

The knowledge that the persona possesses is as follows: \textbf{[knowledge]}
\\
  \hline
\end{tabularx}
\caption{\label{table_B5}
The designed prompt for generating posts content based on world knowledge. The word ``topic output'' in square brackets represents a given topic, the word ``persona'' in square brackets represents the persona information of the agent, and the word ``knowledge'' in square brackets represents personalized world knowledge.
}
\end{table*}

\begin{table*}[htp]
\begin{tabularx}{1.0\textwidth} {
  >{\hsize=1\hsize\linewidth=\hsize}X
}
  \hline
Assume you are the person described in [Persona information], and you usually browse social media and regularly post. Please generate a post suitable for posting on Twitter based on the provided topic. Directly output the generated post content, and do not insert images or videos. The generated posts should not be a duplication of persona information, and the post content should be about a single topic and be specific and rich. The length of the post must be limited to 500 characters. The post topic and persona information are as follows:

The post topic is: \textbf{[topic output]}

Persona information (JSON format) is as follows: \textbf{[persona]}
\\
  \hline
\end{tabularx}
\caption{\label{table_B6}
The designed prompt for generating posts content without world knowledge. The word ``topic output'' in square brackets represents a given topic, and the word ``persona'' in square brackets represents the persona information of the agent.
}
\end{table*}

\begin{table*}[htp]
\begin{tabularx}{1.0\textwidth} {
  >{\hsize=1\hsize\linewidth=\hsize}X
}
  \hline
Assume that you are the person described in [Persona information], you usually browse social media and perform social behaviors such as liking and posting, and you have an activity level of: \textbf{[active prob]} full activity level is 1). In order to be consistent with your persona\'s behavior, You need to plan and schedule your behavior and generate a coarse-grained planning table containing the frequency of the behavior and the duration of the behavior, all using a 24-hour time frame and providing only one time period for browsing and posting. Please strictly follow the following examples to generate a plan. Here is an example, please follow the format in the example for the output: 

Browsing time period: xx:xx-xx:xx 

Probability of liking: x\% 

Probability of forwarding: x\% 

Probability of commenting: x\% 

Posting time period: day x-xx:xx-xx:xx Frequency of posting: x times per week 

Your persona information is as follows:

Persona information: \textbf{[persona]}
\\
  \hline
\end{tabularx}
\caption{\label{table_C1}
The designed prompt for generating the plan of the action. The word ``active prob'' in square brackets represents the user active of the agent, and the word ``persona'' in square brackets represents the persona information of the agent.
}
\end{table*}

\clearpage

\begin{table*}[htp]
\begin{tabularx}{1.0\textwidth} {
  >{\hsize=1\hsize\linewidth=\hsize}X
}
  \hline
Assume you are the person described in [Persona information], when you browse social media, you like, repost, and comment on multiple posts based on how much you like them. Please reflect and think based on your historical behavior and think about which user you want to follow. Please strictly enter the user ID you want to follow or ``do not follow'', no other content is required. Your persona information and historical behaviors are as follows:

Persona information: \textbf{[persona]} 

The content of multiple posts and your operations are as follows: \textbf{[reflect content]}
\\
  \hline
\end{tabularx}
\caption{\label{table_D1}
The designed prompt for the following user. The word ``persona'' in square brackets represents the persona information of the agent, and the word ``reflect content'' in square brackets represents the historical action information of the agent.
}
\end{table*}

\begin{table*}[htp]
\begin{tabularx}{1.0\textwidth} {
  >{\hsize=1\hsize\linewidth=\hsize}X
}
  \hline
The \textbf{[k]}-th post was posted by user \textbf{[uid]}, and the content of the post is: \textbf{[post summary]}. Your action on this post is: \textbf{[action]}.

+

The \textbf{[k+1]}-th post was posted by user \textbf{[uid]}, and the content of the post is: \textbf{[post summary]}. Your action on this post is: \textbf{[action]}.

+

The \textbf{[k+2]}-th post was posted by user \textbf{[uid]}, and the content of the post is: \textbf{[post summary]}. Your action on this post is: \textbf{[action]}.

+

...
\\
  \hline
\end{tabularx}
\caption{\label{table_D2}
The designed prompt for splicing historical action information. The ``k'' in square brackets is the serial number of the post, the ``uid'' in square brackets is the ID of the post author, and the word ``post summary'' in square brackets is a refinement of the post content, and the word ``action'' in square brackets represents the information about the current action.
}
\end{table*}

\begin{table*}[htp]
\begin{tabularx}{1.0\textwidth} {
  >{\hsize=1\hsize\linewidth=\hsize}X
}
  \hline
The following is a post from social media, please generate a concise summary of no more than 50 words. The content of the post is as follows: \textbf{[post]}
\\
  \hline
\end{tabularx}
\caption{\label{table_E1}
The designed prompt for summarizing the post content. The word ``post'' in square brackets represents the content of the post.
}
\end{table*}

\clearpage

\begin{table*}[htp]
\centering
\scalebox{0.8}{
\begin{tabular}{lcccccccccc}
\toprule
& \textbf{\multirow{2}{*}{Method}} &  \textbf{\multirow{2}{*}{Action}} & \multicolumn{6}{c}{\textbf{Automatic}}  & & \multicolumn{1}{c}{\textbf{Human}} \\
\cline{4-9} \cline{11-11}
\textbf{} & & & BERTScore & $\Delta BS\uparrow$ & C.score & $\Delta C\uparrow$ & Dis-1$\uparrow$ & Dis-2$\uparrow$ & & Rationality$\uparrow$ \\
\hline
\textbf{\multirow{14}{*}{First Stage}} & \multirow{2}{*}{Ours} & not like & 0.433 & \multirow{2}{*}{0.007} & 0.934 & \textbf{\multirow{2}{*}{0.038}} & \multirow{2}{*}{-} & \multirow{2}{*}{-} &  & \textbf{\multirow{2}{*}{0.678}} \\
& & like & 0.440 & & 0.972 & & & & \\
\cline{3-11}
 & \multirow{2}{*}{$w/o$ PD} & not like & 0.429 & \textbf{\multirow{2}{*}{0.012}} & 0.948 & \multirow{2}{*}{0.008} & \multirow{2}{*}{-} & \multirow{2}{*}{-} &  &  \multirow{2}{*}{0.667} \\
 & & like & 0.441 & & 0.956 &  &  & &\\
 
 \cline{2-11}
  & \multirow{2}{*}{Ours} & not reblog & 0.432 & \textbf{\multirow{2}{*}{0.006}} & 0.928 & \textbf{\multirow{2}{*}{0.039}} & \multirow{2}{*}{-} & \multirow{2}{*}{-} &  &  \textbf{\multirow{2}{*}{0.656}} \\
 & & reblog & 0.438 & & 0.967 &  &  & &\\
 \cline{3-11}
  & \multirow{2}{*}{$w/o$ PD} & not reblog & 0.429 & \multirow{2}{*}{0.006} & 0.949 & \multirow{2}{*}{0.000} & \multirow{2}{*}{-} & \multirow{2}{*}{-}  & & \multirow{2}{*}{0.644} \\
 & & reblog & 0.435 & & 0.949 &  &  & &\\
 \cline{2-11}
 
 & \multirow{2}{*}{Ours} & not comment & 0.433 & \multirow{2}{*}{0.002} & 0.946 & \multirow{2}{*}{0.002} & - & - &  & \textbf{\multirow{2}{*}{0.544}}  \\
 & & comment & 0.435 & & 0.944 & & 3.38 & 19.05 & &  \\
  \cline{3-11}
  & \multirow{2}{*}{$w/o$ PD} & not comment & 0.426 & \textbf{\multirow{2}{*}{0.005}} &  0.933 &  \textbf{\multirow{2}{*}{0.023}} & - & - &  &  \multirow{2}{*}{0.533} \\
 & & comment & 0.431 & & 0.956 & & \textbf{4.24} & \textbf{22.24} & &\\
 \cline{2-11}
 
& Ours & post & 0.465 & - & 0.973 & - & \textbf{22.40} & \textbf{63.88} & & - \\
\cline{3-11}
& $w/o$ PD & post & 0.472 & - & 0.979 & - & 21.25 & 62.14 & & - \\
\hline

\textbf{\multirow{14}{*}{Second Stage}} & \multirow{2}{*}{Ours} & not like & 0.428 & \multirow{2}{*}{0.007} & 0.931  & \textbf{\multirow{2}{*}{0.022}} & \multirow{2}{*}{-} & \multirow{2}{*}{-} & & \textbf{\multirow{2}{*}{0.778}} \\
& & like & 0.435 & & 0.953 &  &  & & \\
\cline{3-11}
& \multirow{2}{*}{$w/o$ PD} & not like & 0.434 & \textbf{\multirow{2}{*}{0.011}} & 0.975 & \multirow{2}{*}{0.006} & \multirow{2}{*}{-} & \multirow{2}{*}{-} &  &  \multirow{2}{*}{0.622} \\
& & like & 0.445 & & 0.981 & &  &  & &\\

\cline{2-11}
& \multirow{2}{*}{Ours} & not reblog & 0.428  & \textbf{\multirow{2}{*}{0.006}} & 0.937 & \multirow{2}{*}{0.005} & \multirow{2}{*}{-} & \multirow{2}{*}{-} & & \textbf{\multirow{2}{*}{0.622}} \\
 & & reblog & 0.434 & & 0.942 & &  &  & &  \\
\cline{3-11}
& \multirow{2}{*}{$w/o$ PD} & not reblog & 0.433 & \multirow{2}{*}{0.006} & 0.974 & \textbf{\multirow{2}{*}{0.010}} & \multirow{2}{*}{-} & \multirow{2}{*}{-} &  &  \multirow{2}{*}{0.622} \\
& & reblog & 0.439 & & 0.984 &  &  & &\\
 
 \cline{2-11}
& \multirow{2}{*}{Ours} & not comment & 0.428  & \multirow{2}{*}{0.003} & 0.962 & \multirow{2}{*}{-0.023} & - & - & &  \textbf{\multirow{2}{*}{0.477}} \\
 & & comment & 0.431 & & 0.939 & & 2.98 & 17.17 & &   \\
 \cline{3-11}
 & \multirow{2}{*}{$w/o$ PD} & not comment & 0.426 & \textbf{\multirow{2}{*}{0.009}} & 0.879 &  \textbf{\multirow{2}{*}{0.100}} & - & - &  &  \multirow{2}{*}{0.433} \\
& & comment & 0.435 & & 0.979 & & \textbf{3.51} & \textbf{20.15} & &\\
 \cline{2-11}

 & Ours & post & 0.466 & - & 0.983 & - & \textbf{22.24} & \textbf{64.10} & & - \\
 \cline{3-11}
 & $w/o$ PD & post & 0.470 & -  & 0.979 & - & 21.40 & 62.19 & & - \\
 \cline{3-11}
 & $w/o$ KB & post & 0.472 & - & 0.981 & - & 20.55 & 61.49 & & - \\
\bottomrule
\end{tabular}
}
\caption{\label{table-Ab1}
Ablation results of agent actions.}
\end{table*}


\begin{table*}[htp]
\centering
\scalebox{0.95}{
\begin{tabular}{lcccccc}
\toprule
\textbf{} & & & \textbf{Info.$\uparrow$}  & \textbf{Cons.$\uparrow$}  & \textbf{Hum.$\uparrow$}  & \textbf{Know.$\uparrow$} \\
\hline
\textbf{\multirow{4}{*}{Comment}} & \multirow{2}{*}{Ours} & first stage & 2.422 & 1.922 & 2.533 & - \\
& & second stage & 2.522 & 2.011 & 2.600 & - \\
\cline{2-7}
& \multirow{2}{*}{$w/o$ PD} & first stage & 2.244 & 1.911 & 2.011 & - \\
& & second stage & 1.856 & 2.056 & 1.967 & - \\
\hline

\textbf{\multirow{5}{*}{Post}} & \multirow{2}{*}{Ours} & first stage & 2.733 & 2.911 & 2.600 & 0.889  \\
& & second stage  & 2.589 & 2.944 & 2.667 & 0.778  \\
\cline{2-7}

& \multirow{2}{*}{$w/o$ PD} & first stage & 2.456 & 2.756 & 2.444 & 0.889 \\
& & second stage & 2.411 & 2.833 & 2.500 & 0.889 \\
\cline{2-7}
& $w/o$ KB & - & 2.233 & 2.789 & 2.444 & - \\
\bottomrule
\end{tabular}
}
\caption{\label{table-Ab2}
Human evaluation of ablation results for text generation quality.} 
\end{table*}

\clearpage

\begin{table*}[htp]
\centering
\scalebox{0.99}{
\begin{tabularx}{1.0\textwidth} {
  >{\hsize=.25\hsize\linewidth=\hsize}X
  >{\hsize=1.75\hsize\linewidth=\hsize}X
}
      \hline
      Persona

       & \{\textbf{"name"}: "Sarah", 
       \textbf{"age"}: 24, 
       \textbf{"gender"}: "Female", 
       
       \textbf{"nationality"}: "Canadian", 
       \textbf{"personality"}: "Compassionate, Dedicated", 
       
       \textbf{"hobbies"}: "Volunteering at animal shelters, Running, Reading", 
       
       \textbf{"detailed historical behaviour information"}: "Sarah has always had a deep love for animals, especially dogs. She started volunteering at local animal shelters during her high school years and continued to do so throughout her undergraduate studies. Sarah believes that dogs bring immense joy and comfort to people's lives and she actively promotes responsible pet ownership. She spends her free time training and socializing the shelter dogs to increase their chances of finding forever homes. Sarah is also an avid runner and uses her passion for running to raise funds for animal welfare organizations. She participates in marathons and has successfully completed several races. Additionally, she enjoys reading about dog behavior and training techniques to further enhance her knowledge and skills in working with dogs.", 
       
       \textbf{"preferences for social media content"}: "On social media platforms, Sarah loves sharing heartwarming stories and pictures of the dogs she has helped rescue and rehabilitate. She often posts updates on their progress, including their successful adoptions. Sarah also shares her running journey, documenting her training regimen and the races she participates in. She hopes to inspire others to use their hobbies and passions to make a positive impact. She is open to recommendations and discussions about dog training, running techniques, and books about animal behavior.", 
       
       \textbf{"knowledge"}: "Sarah has gained extensive knowledge about dog behavior and training throughout her years of volunteering and personal research. She is familiar with various training methods, including positive reinforcement and clicker training. Sarah understands the importance of socialization and mental stimulation for dogs and actively incorporates these practices in her volunteering efforts. She keeps up with the latest advancements in animal psychology and training techniques. In terms of running, Sarah is knowledgeable about proper running form, injury prevention, and training plans for different distances. She stays updated on the latest trends in running gear and technology. Sarah also has a good grasp of animal welfare laws and regulations in Canada and is well-informed about current issues in the field."\}
  \\
      \hline
      Retrieved
      
      persona

      information
      & Name: Sarah, age: 24, gender: Female, nationality: Canadian, 
      
       Personality: Compassionate, Dedicated, 
       
       Hobbies: Volunteering at animal shelters, Running, Reading, 
      
      Detailed historical behaviour information: Additionally, she enjoys reading about dog behavior and training techniques to further enhance her knowledge and skills in working with dogs.
      
      Preferences for social media content: She is open to recommendations and discussions about dog training, running techniques, and books about animal behavior.
      
      Knowledge: Sarah has gained extensive knowledge about dog behavior and training throughout her years of volunteering and personal research.
\\

\hline
      Retrieved 
      
      personalized
      
      knowledge & Title: Paul Owens (dog trainer) 
      
      Text: Paul Owens is the author of the bestselling dog training book, ``The Dog Whisperer, Beginning and Intermediate Training for Puppies and Dogs'' (1999; 2nd edition 2007). His newest puppy training book is ``The Puppy Whisperer, A Compassionate, Nonviolent Guide to Early Training and Care'' (2007).
\\

      \hline

      \textbf{Generated}
      
      \textbf{post} & Looking for insights into dog behavior and effective training techniques? Dive into Paul Owens' ``The Dog Whisperer'' series! From puppy training to understanding your furry friend's needs, these books offer compassionate, nonviolent methods. Enhance your bond with your canine companion! \#DogTraining \#BookRecommendations
\textit{}
\\
      \hline
    \end{tabularx}
    }
    \caption{\label{tableG} 
    An example of the post action of the agent.}
\end{table*}

\begin{table*}[htp]
\centering
\scalebox{0.95}{
\begin{tabularx}{1.0\textwidth} {
  >{\hsize=.25\hsize\linewidth=\hsize}X
  >{\hsize=1.75\hsize\linewidth=\hsize}X
}
      \hline
      Persona

       & \{\textbf{"name"}: "Sarah", 
       \textbf{"age"}: 24, 
       \textbf{"gender"}: "Female", 
       
       \textbf{"nationality"}: "Canadian", 
       \textbf{"personality"}: "Compassionate, Dedicated", 
       
       \textbf{"hobbies"}: "Volunteering at animal shelters, Running, Reading", 
       
       \textbf{"detailed historical behaviour information"}: "Sarah has always had a deep love for animals, especially dogs. She started volunteering at local animal shelters during her high school years and continued to do so throughout her undergraduate studies. Sarah believes that dogs bring immense joy and comfort to people's lives and she actively promotes responsible pet ownership. She spends her free time training and socializing the shelter dogs to increase their chances of finding forever homes. Sarah is also an avid runner and uses her passion for running to raise funds for animal welfare organizations. She participates in marathons and has successfully completed several races. Additionally, she enjoys reading about dog behavior and training techniques to further enhance her knowledge and skills in working with dogs.", 
       
       \textbf{"preferences for social media content"}: "On social media platforms, Sarah loves sharing heartwarming stories and pictures of the dogs she has helped rescue and rehabilitate. She often posts updates on their progress, including their successful adoptions. Sarah also shares her running journey, documenting her training regimen and the races she participates in. She hopes to inspire others to use their hobbies and passions to make a positive impact. She is open to recommendations and discussions about dog training, running techniques, and books about animal behavior.", 
       
       \textbf{"knowledge"}: "Sarah has gained extensive knowledge about dog behavior and training throughout her years of volunteering and personal research. She is familiar with various training methods, including positive reinforcement and clicker training. Sarah understands the importance of socialization and mental stimulation for dogs and actively incorporates these practices in her volunteering efforts. She keeps up with the latest advancements in animal psychology and training techniques. In terms of running, Sarah is knowledgeable about proper running form, injury prevention, and training plans for different distances. She stays updated on the latest trends in running gear and technology. Sarah also has a good grasp of animal welfare laws and regulations in Canada and is well-informed about current issues in the field."\}
 \\
 \hline
      Browsed
      
      post
      & Diving into the latest MMA showdown! Analyzing jaw-dropping techniques, strategic brilliance, and fighter performances. From ground game to striking precision, let's break it down together! Who impressed you the most? Share your thoughts! \#MMAAnalysis \#FightNightInsights 
\\
\hline
      Retrieved
      
      persona

      information
      & Name: Sarah, age: 24, gender: Female, nationality: Canadian, 
      
      Personality: Compassionate, Dedicated, 
       
      Hobbies: Volunteering at animal shelters, Running, Reading, 
       
      Detailed historical behaviour information: Additionally, she enjoys reading about dog behavior and training techniques to further enhance her knowledge and skills in working with dogs.
      
      Preferences for social media content: She is open to recommendations and discussions about dog training, running techniques, and books about animal behavior.
      
      Knowledge: Sarah has gained extensive knowledge about dog behavior and training throughout her years of volunteering and personal research.
\\

\hline
      Like & The agent chooses to like this post.
\\

\hline
      Reblog & The agent chooses to reblog this post.
\\

      \hline

      Comment &
      The agent chooses to comment on this post.

      Comment content: Wow, another incredible MMA showdown! While I'm usually more focused on animal welfare, it's fascinating to see the dedication and skill these fighters bring to the ring. Each match is like a carefully choreographed dance of athleticism and strategy. Kudos to all the fighters for their hard work and determination! \#MMAAnalysis \#FightNightInsights
\textit{}
\\
      \hline
    \end{tabularx}
    }
    \caption{\label{tableG2} 
    An example of like, reblog, and comment actions of the agent.}
\end{table*}

\end{document}